\documentclass{article}
\usepackage{spconf,amsmath,amssymb,graphicx}

\usepackage{multirow}
\usepackage{cite}
\RequirePackage{booktabs}
\usepackage{bbding}  
\usepackage{CJKutf8}  

\newcommand\blfootnote[1]{%
  \begingroup
  \renewcommand\thefootnote{}\footnote{#1}%
  \addtocounter{footnote}{-1}%
  \endgroup
}

\title{Non-autoregressive Mandarin-English\\Code-switching Speech Recognition}
\name{Shun-Po Chuang$^{\dagger}$, Heng-Jui Chang$^{\dagger}$, Sung-Feng Huang, Hung-yi Lee}
\address{College of Electrical Engineering and Computer Science, National Taiwan University}
\begin{document}
%
\maketitle
\begin{abstract}
Mandarin-English code-switching (CS) is frequently used among East and Southeast Asian people.
However, the intra-sentence language switching of the two very different languages makes recognizing CS speech challenging.
Meanwhile, the recent successful non-autoregressive (NAR) ASR models remove the need for left-to-right beam decoding in autoregressive (AR) models and achieved outstanding performance and fast inference speed, but it has not been applied to Mandarin-English CS speech recognition.
This paper takes advantage of the Mask-CTC NAR ASR framework to tackle the CS speech recognition issue.
We further propose to change the Mandarin output target of the encoder to Pinyin for faster encoder training and introduce the Pinyin-to-Mandarin decoder to learn contextualized information.
Moreover, we use word embedding label smoothing to regularize the decoder with contextualized information and projection matrix regularization to bridge that gap between the encoder and decoder.
We evaluate these methods on the SEAME corpus and achieved exciting results.
\end{abstract}
\blfootnote{$^\dagger$ Equal contribution.}
\blfootnote{We thank National Center for High-performance Computing (NCHC) of National Applied Research Laboratories (NARLabs) in Taiwan for providing computational and storage resources.
}

\section{Introduction}
\label{sec:intro}

Code-switching (CS) is a phenomenon of using two or more languages within a sentence in text or speech.
This practice is common in regions worldwide, especially in East and Southeast Asia, where people frequently use Mandarin-English CS speech in daily conversations.
Thus, developing technologies to recognize this type of speech is critical.
Although humans can easily recognize CS speech, automatic speech recognition (ASR) technologies nowadays perform poorly since these systems are primarily trained with monolingual data.
Moreover, very few CS speech corpora are publicly available \cite{lyu2010seame, shen2011cecos, li2012mandarin, baumann2019spoken}, making training high-performance ASR models more challenging for CS speech.

Various approaches were studied to tackle the CS speech recognition problem, including language identity recognition~\cite{zhang2021rnn, li2019towards, shan2019investigating, zeng2019end} and data-augmentation~\cite{du2021data, long2020acoustic, sharma2020improving,Chang2019}.
Many prior works exploited the powerful end-to-end ASR technologies like listen, attend and spell~\cite{chan2016las} and RNN transducers~\cite{graves2012rnnt}.
However, these methods mostly require autoregressive (AR) left-to-right beam decoding, leading to longer processing time and higher latency.
Some studies also demonstrated that processing each language with its encoder or decoder obtained better performance~\cite{zhou2020multi, dalmia2021transformer, lu2020bi}.
This multi-encoder-decoder architecture and AR decoding require significantly more computation time, thus less feasible for real-world applications.
Therefore, in this paper, we leverage non-autoregressive (NAR) ASR technology to tackle this issue.

Unlike the successful AR ASR models~\cite{gulati2020conformer, liu2020improving}, NAR ASR eliminates the left-to-right dependency by directly predicting the whole sequence at once or refine the output sequence in a constant amount of iterations~\cite{chi2020align, chen2019listen, chan2020imputer}.
NAR ASR can thus exploit parallel computing technologies for faster inference.
This paper primarily adopts Mask-CTC~\cite{higuchi2020maskctc, higuchi2021improved-maskctc}, a NAR ASR framework with a NAR conditional masked language model (CMLM) \cite{ghazvininejad-etal-2019-mask} for iterative decoding, possessing competing performance with AR ASR baselines.

Although Mask-CTC provided exciting results in numerous ASR tasks, recognizing CS speech remains challenging.
The CS task suffers from data scarcity~\cite{Sharma2020,chuang2020training}, using deep learning-based methods with millions of parameters like transformer easily lead to overfitting.
Thereby, we use word embedding label smoothing \cite{song2021word} to incorporate additional textual knowledge, bringing more precise semantic and contextual information for better regularization.
Second, the current Mask-CTC framework propagates decoder errors to the encoder, but we wish to increase the connection between the two modules for better performance.
We suggest constraining the encoder's output projection layer similar to the input embedding layer of the decoder to bridge the gap between the encoder and decoder.
Furthermore, we propose to change the encoder's output target from Chinese characters to Pinyin symbols \cite{chan2016online}.
Each Chinese character can be mapped to at least one Pinyin symbol, representing its pronunciation.
Using Pinyin symbols as the target allows the encoder to focus on acoustic modeling and reduces the vocabulary size.

Before this paper, Naowarat et al.~\cite{naowarat2020reducing} used contextualized CTC to force NAR ASR to learn contextual information in CS speech recognition.
In contrast, we utilize CMLM to better modeling linguistic information.
Also, Huang et al.~\cite{huang2021context} developed context-dependent label smoothing methods by removing impossible labels given the previous output token, while our word embedding label smoothing is more straightforward and leverages knowledge from the additional text data.
Song et al. \cite{song2021word} proposed word similarity-based label smoothing very similar to our work.
Instead, they applied this method to monolingual RNN-LM.

We conducted extensive experiments on the SEAME CS corpus~\cite{lyu2010seame} to evaluate our methods.
Next, we showed that the proposed Pinyin-based Mask-CTC and the regularization methods benefit CS speech recognition and offer comparable performance to the AR baseline.
Moreover, error analyses prove our methods useful and provided insights to future improvements of CS speech recognition.

\section{Method}
\label{sec:method}

\subsection{Mask-CTC}
\label{subsec:method}

The Mask-CTC \cite{higuchi2020maskctc, higuchi2021improved-maskctc} model is a NAR ASR framework that adopts the transformer encoder-decoder structure, but the decoder is a CMLM for NAR decoding.

First, in the training phase, given transcribed audio-text pair $(X, Y)$, the acoustic feature sequence $X$ is first encoded with a conformer encoder model~\cite{gulati2020conformer}, and linearly projected to probability distributions over all possible vocabularies $\mathcal{V}$ to minimize the CTC loss~\cite{graves2006ctc}.
Then, some tokens $Y_{\mathrm{mask}} \subset Y$ are randomly replaced with the special token \texttt{<mask>}.
The decoder is trained to predict the masked tokens conditioned on the observed tokens $Y_{\mathrm{obs}} = Y \backslash Y_{\mathrm{mask}}$ and the encoder output.
The Mask-CTC model is thus trained to maximize the log-likelihood
\begin{equation}
    \begin{split}
        \log P_{\mathrm{NAR}}(Y | X) = & ~\alpha \log P_{\mathrm{CTC}}(Y | X) \\
        & + (1 - \alpha) \log P_{\mathrm{CMLM}}(Y_{\mathrm{mask}} | Y_{\mathrm{obs}}, X),
    \end{split}
\end{equation}
where $0 \leq \alpha \leq 1$ is a tunable hyper-parameter.

At the decoding stage, the encoder output sequence is first transformed by CTC greedy decoding, denoted as $\hat{Y}$.
Next, tokens with probability lower than a specified threshold $P_{\mathrm{thres}}$ in $\hat{Y}$ are masked with \texttt{<mask>} for the CMLM to predict, denoted as $\hat{Y}_{\mathrm{mask}}$.
Then, the masked tokens are gradually recovered by the CMLM in $K$ iterations.
In the $k^{\mathrm{th}}$ iteration, the most $\lfloor |\hat{Y}_{\mathrm{mask}}| / K \rfloor$ probable tokens in the masked sequence are recovered by calculating
\begin{equation}
    y_u = \underset{y \in \mathcal{V}}{\arg \max} ~P_{\mathrm{CMLM}}(y_u = y | \hat{Y}^{(k)}_{\mathrm{obs}}, X),
\end{equation}
where $y_u$ is the $u^{\mathrm{th}}$ output token and $\hat{Y}^{(k)}_{\mathrm{obs}}$ is the sequence including the unmasked tokens and the masked tokens recovered in the first $(k - 1)$ iterations.
The Mask-CTC ASR framework achieves performance close to AR ASR models but with significantly faster decoding and lower latency \cite{higuchi2020maskctc, higuchi2021improved-maskctc}.

\subsection{Pinyin as Output Target}
\begin{figure}[t]
	\centering
	\includegraphics[width=\linewidth]{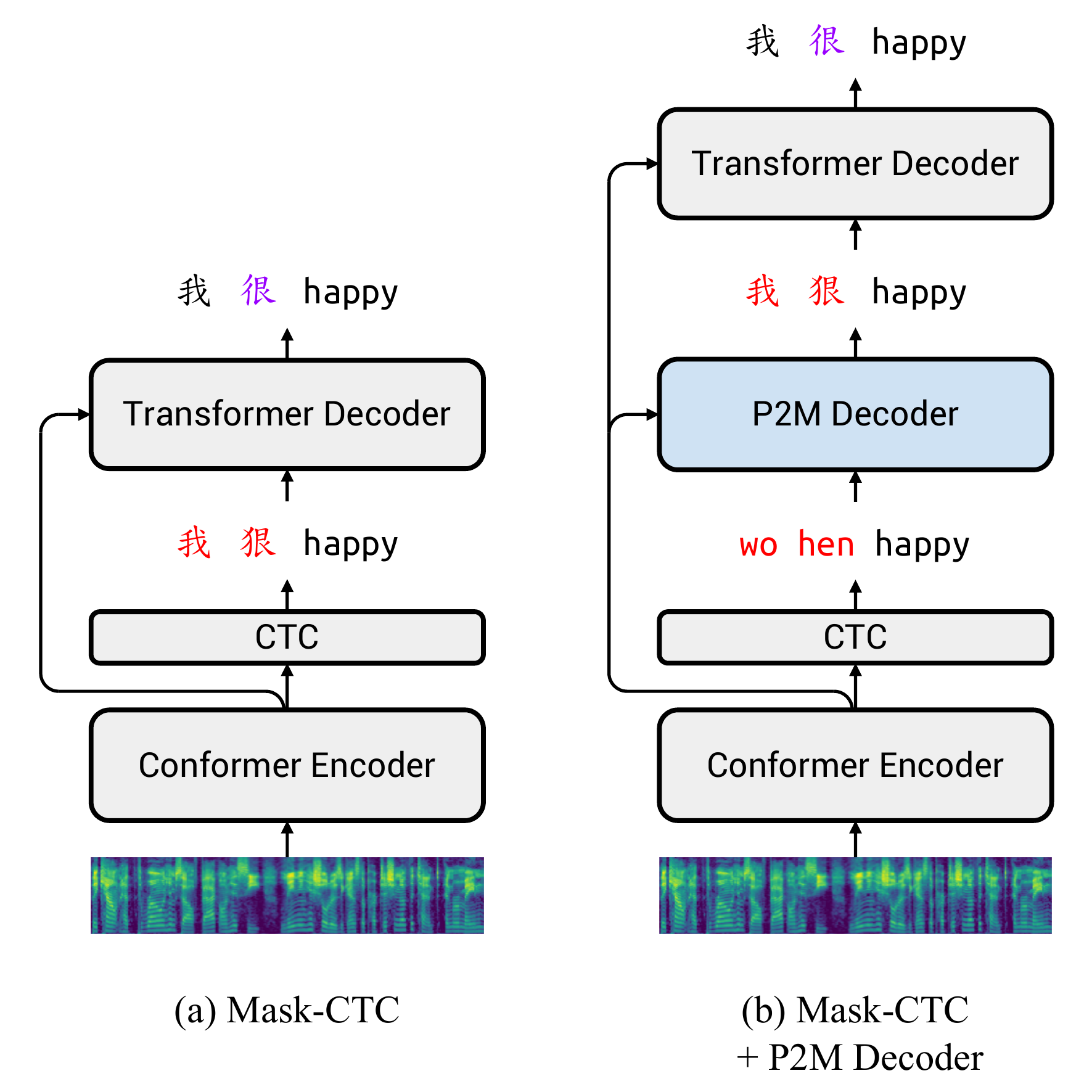}
	\caption{
	(a) The original Mask-CTC \cite{higuchi2020maskctc} framework and (b) the proposed Mask-CTC + Pinyin-to-Mandarin decoder framework.
	The example "\begin{CJK*}{UTF8}{bkai}我(wo)很(hen)\end{CJK*} happy" means "I am very happy."
	However, the character "\begin{CJK*}{UTF8}{bkai}很(hen)\end{CJK*}" is misspelled as "\begin{CJK*}{UTF8}{bkai}狠(hen)\end{CJK*}" which have the same pronunciation but different meanings.
	The final transformer decoders can recover the wrong character.
	}
	\label{fig:model}
\end{figure}

In this paper, we introduce the Pinyin-to-Mandarin (P2M) decoder for dealing with CS data, as shown in Fig.~\ref{fig:model}.
One of the challenges of training Mandarin-English CS speech recognition models is that jointly modeling the two very different languages is tricky.
Roughly 5k characters are frequently used in Mandarin, consisting of approximately 400 pronunciations or 1500 with tones involved.
Thereby having many characters with identical pronunciation, increasing ambiguity and making ASR models prone to predict incorrect characters with the correct pronunciations.
Moreover, a large vocabulary size is difficult for CTC training.
We thus propose changing the encoder's targets from Chinese character to Pinyin to separate the process of acoustic and language modeling.

Each Chinese character can be mapped to at least one Pinyin symbol, representing the pronunciation, similar to phoneme representations.
Replacing characters with Pinyin significantly reduces vocabulary size and allows the encoder to focus on learning acoustic modeling.
The Pinyin system's pronunciation rules are similar to English since Pinyin symbols can be represented in Latin letters, providing a more intuitive way to utilize a pre-trained English ASR model for initialization.

\noindent\textbf{P2M Decoder.}
Next, we propose a P2M decoder to map Pinyin back to Chinese characters using a single-layer decoder, as shown in the middle of Fig.~\ref{fig:model}.
The P2M model is trained with Pinyin-Mandarin character sequence pairs conditioned on the encoder's output.
Based on acoustic-contextualized information, the P2M model can translate Pinyin-English mixed sequence to Mandarin-English because typical phonetic sequences in Mandarin are usually different from English.
The original Mask-CTC uses Mandarin-English sentences to train the decoder; this approach further applies the triplet pair (Pinyin, Mandarin, English) for training, providing more information to train the final Mandarin-English decoder.
We encourage the P2M model to learn a Pinyin-to-Mandarin character translation with a randomly masking technique.

\subsection{Word Embedding Label Smoothing Regularization}
\label{subsec:wemb_ls}
In this section, we introduce a label smoothing~\cite{szegedy2016rethinking} method using pre-trained word embedding.
As mentioned earlier, the scarcity of transcribed CS data makes CS ASR models easily overfit and unable to generalize.
Hence, to improve the performance and prevent the CMLM decoder from overconfidence, we wish to regularize the model parameters and bring more textual knowledge via label smoothing.

Conventionally, label smoothing reduces the ground truth label's probability to $1 - \epsilon$ for a small constant $0 < \epsilon < 1$, while all the other labels are equally assigned with a small constant probability.
Although label smoothing is useful in regularization, it is incapable of exploiting the target's meaning and similarity with other targets.
Therefore, we distill knowledge from semantic-rich word embedding pre-trained from a large amount of text data.
Word embedding reflects semantic-similar and contextual-relevant words that would have similar word representations.
By leveraging such properties would allow the model to learn the semantic-contextual information implicitly.

In this paper, we determine the possible labels by calculating the cosine similarity between word embedding.
The original method proposed by Song et al. \cite{song2021word} chooses similar words with cosine similarity higher than a threshold $\tau$.
For a ground truth label $\hat{y}$, the set of similar words are chosen as
\begin{equation}
    D = \left\{ y \in \mathcal{V} \backslash \{\hat{y}\}~\left|~ \cos \left( \mathbf{e}(\hat{y}), \mathbf{e}(y) \right) \geq \tau \right.\right\},
\end{equation}
where $\mathbf{e}$ denotes pre-trained word embedding and $\cos(\cdot, \cdot)$ returns the cosine similarity between the two inputs.
However, determining the best value of $\tau$ for each dataset requires multiple experiments, which is problematic since training an ASR model is time-consuming and thus unsuitable in our case.
Moreover, datasets with different languages, vocabulary sizes, or word embedding types should set different thresholds to find a proper $D$ for each token.
To mitigate this issue, we propose setting the size of $D$ to a fixed value $N$.
In this case, we guarantee top $N$ similar words can be collected for each token in the vocabulary.
The set of similar words can thus be obtained by
\begin{equation}
    D = 
    \underset{y \in \mathcal{V} \backslash \{\hat{y}\}}{\arg\mathrm{top}N} ~ 
    \cos \left( \mathbf{e}(\hat{y}), \mathbf{e}(y) \right),
\end{equation}
where the $\arg\mathrm{top}N$ operator returns a set of $N$ indices indicating the labels with the top $N$ highest scores.
The probabilities of each label can be written as
\begin{equation}
    P(y) = 
    \left\{ \begin{array}{rl}
        1 - \epsilon, & y = \hat{y} \\
        \frac{\epsilon}{N}, & y \in D \\
        0, & \mathrm{otherwise}
    \end{array}\right. .
\end{equation}

We expect the label smoothing method provides a better regularization effect.
Moreover, the word embeddings can be trained with additional text data other than CS data, introducing richer information to the decoder.

\subsection{Projection Matrix Regularization}
\label{subsec:proj_mat}
Here, we propose regularizing the projection matrices in the ASR model to bridge the gap between the encoder and decoder.
The Mask-CTC model's components are trained jointly, and the errors of the decoder are backpropagated through the encoder.
During training, the decoder takes masked ground-truth labels as its input; nevertheless, during inference, the decoder's input is the erroneous output of the encoder, making it more challenging to decode correctly.
This mismatch might lead to degraded performance of the ASR.
Thus, we propose to mitigate this problem by making the encoder output projection matrix and the decoder input embedding matrix similar to each other.

Another purpose of this method is to fuse the knowledge in the embedding layers.
The encoder mainly focuses on acoustic modeling, making the output embeddings contain more phonetic information and easily output wrong tokens with similar pronunciation.
In contrast, the decoder input embedding contains linguistic information, regularizing these embedding spaces makes the decoder input embedding involve more phonetic information.
Therefore, the CMLM can learn to predict correct sentences based on the phonetic-linguistic mixed information.

The computation of the projection matrix regularization loss is shown as follows.
The last layer of the encoder is a matrix $\mathbf{W}_{\mathrm{CTC}} \in \mathbb{R}^{d \times |\mathcal{V}|}$ for linearly projecting encoded features of dimension $d$ to a probability distribution over all $|\mathcal{V}|$ possible output labels.
The first layer of the decoder is the embedding layer that transforms one-hot vectors representing text tokens to continuous hidden features with a matrix $\mathbf{W}_{\mathrm{emb}}^T \in \mathbb{R}^{d \times |\mathcal{V}|}$.
Matrices $\mathbf{W}_{\mathrm{CTC}}$ and $\mathbf{W}_{\mathrm{emb}}^T$ can be respectively written as $[\boldsymbol{w}_{\mathrm{CTC}, 1} ~\dots~ \boldsymbol{w}_{\mathrm{CTC}, |\mathcal{V}|}]$ and $[\boldsymbol{w}_{\mathrm{emb}, 1} ~\dots~ \boldsymbol{w}_{\mathrm{emb}, |\mathcal{V}|}]$, where $\boldsymbol{w}$ are column vectors of dimension $d$.
Here we wish the two matrices have similar behavior, and thus apply cosine embedding loss to constrain the two matrices as
\begin{equation}
    \mathcal{L}_{\mathrm{MatReg}} = 
    \frac{1}{|\mathcal{V}|} \sum_{v=1}^{|\mathcal{V}|}
    \left[ 1 - \cos\left( \boldsymbol{w}_{\mathrm{CTC}, v}, \boldsymbol{w}_{\mathrm{emb}, v} \right) \right].
\end{equation}
This loss function can also be used to build a relation between the output layer of the P2M decoder and the input embedding layer of CMLM.
Additionally, an alternative solution is to share the same weights in $\mathbf{W}_{\mathrm{CTC}}$ and $\mathbf{W}_{\mathrm{emb}}^T$; however, we found this solution severely damaged the performance since the two matrices still have some different functionalities.

The combined loss function is
\begin{equation}
    \label{eq:tot_loss}
    \begin{split}
        \mathcal{L} = & -\alpha\log P_{\mathrm{CTC}}(Y^{\mathrm{pyin}}|X)  \\
        & -(1 - \alpha) \log P_{\mathrm{P2M}}(Y^{\mathrm{char}}_{\mathrm{mask}} | Y^{\mathrm{pyin}}_{\mathrm{obs}}, X) \\
        & -(1 - \alpha) \log P_{\mathrm{CMLM}}(Y^{\mathrm{char}}_{\mathrm{mask}} | Y^{\mathrm{char}}_{\mathrm{obs}}, X) \\
        & + \beta \mathcal{L}_{\mathrm{MatReg}},
    \end{split}
\end{equation}
where $Y^{\mathrm{pyin}}$ and $Y^{\mathrm{char}}$ respectively represent using Pinyin and character as the Mandarin tokens.
In this paper, the log probabilities $P_{\mathrm{P2M}}$ and $P_{\mathrm{CMLM}}$ include either conventional or word embedding label smoothing.

\begin{table}[t]
	\caption{The duration and the composition of Mandarin, English, and code-switching utterances in the training/validation/testing sets of the SEAME corpus.
	}
	\label{tab:corpus}
	\centering
	\vspace{10pt}
	\begin{tabular}{lcccc}
        \toprule
        & \textbf{train} & \textbf{val} & \textbf{devman} & \textbf{devsge} \\
        \midrule
        Duration (hours) &  114.6 & 6.0 & 7.5 & 3.9 \\
        Mandarin & 19.3\% & 19.1\% & 19.9\% & 8.9\% \\
        English & 23.8\% & 24.4\% & 12.4\% & 49.8\% \\
        Code-switching & 56.9\% & 56.5\% & 67.7\% & 41.3\% \\
		\bottomrule
	\end{tabular}

\end{table}

\section{Experiment}

\subsection{Data}

To evaluate the proposed methods for CS speech recognition, we used a CS corpus and three monolingual corpora.

\noindent \textbf{SEAME Corpus}:
SEAME~\cite{lyu2010seame} is a Mandarin-English CS conversational speech corpus collected in Singapore.
This corpus is composed of spontaneous and noisy conversations in interviews.
The two evaluation sets, \textbf{devman} and \textbf{devsge}, are respectively biased toward Mandarin and English.
We excluded all the testing data from the SEAME corpus and used the remaining for the train/validation set.
We used all training data except for the ablation study in Sec. \ref{subsec:ablation}, which only used 10\% of data.
The statistics are listed in Table~\ref{tab:corpus}.
The baseline model used subword units~\cite{kudo-richardson-2018-sentencepiece} for English, Chinese characters for Mandarin, and an additional \texttt{<noise>} token, resulting in a vocabulary size of 5751.
We used 2704 Chinese characters and were reduced to 390 Pinyin tokens for training the P2M decoder.
The \textit{pypinyin} package was used for Pinyin-Mandarin mapping. \footnote{\texttt{https://pypi.org/project/pypinyin/}}

\noindent \textbf{Monolingual Datasets}:
We used the LibriSpeech~\cite{panayotov2015librispeech} English corpus of approximately 960 hours of training data for model pre-training.
Because text data in SEAME are noisy and scarce, we combined SEAME, the TEDLIUM2 English corpus~\cite{rousseau2014tedlium2}, and the AISHELL1 Mandarin corpus~\cite{bu2017aishell} to train a 64-dimensional skip-gram model using the fastText toolkit~\cite{bojanowski2016enriching}.
We chose TEDLIUM2 rather than LibriSpeech for training word embedding since it has shorter and more spontaneous utterances similar to SEAME.

\subsection{Model}

All experiments were based on the ESPnet toolkit \cite{watanabe2018espnet} and followed its LibriSpeech training recipe.
Audio features were extracted to global mean-variance normalized 83-dimensional log-Mel filterbank and pitch features.
Speed perturbation \cite{ko2015speed-perturb} and SpecAugment \cite{park2019specaug} were added throughout the training process for data augmentation.
Conventional label smoothing was applied to all experiments except for those with the word embedding label smoothing technique.
The encoder architecture for both AR and NAR models was a 12-layer conformer encoder with a dimension of 512 and 8 attention heads per layer \cite{gulati2020conformer}.
The P2M decoder was a single-layer transformer decoder, and the CMLM decoder was a 6-layer transformer decoder \cite{vaswani2017attention}, both decoders had a feed-forward dimension of 2048.
The ASR models in our experiments were all initialized with a pre-trained conformer ASR provided by ESPnet.

We evaluated the performance of our models with token error rate (TER) and real-time factor (RTF), where the tokens refer to Chinese characters and English words.
RTF is an indicator for demonstrating the advantage of NAR models' fast inference speed over AR models, which is calculated as dividing the total decoding time by the total audio signal length.
Therefore, lower RTF equals lower computation time and more desirable for real-world applications.
We set the hyper-parameters in Eq. (\ref{eq:tot_loss}) to $\alpha=0.3$ and $\beta=10^{-4}$, and set $\epsilon=0.1$ and $N=10$ for the label smoothing loss.

We found iterative refinement in Mask-CTC provided little improvement in our experiments, probably because the spontaneous and code-switched SEAME corpus.
Compared with the clean and monolingual corpora used in previous studies \cite{higuchi2020maskctc, higuchi2021improved-maskctc}, SEAME made the CMLM decoder of Mask-CTC more challenging to learn.
Therefore, we directly predicted the output with a single pass through the decoder.

\subsection{Pinyin Decoder and Regularization Methods}
\begin{table}[t]
	\caption{
	    TERs(\%) and RTF of the AR and NAR ASR models trained with all data from SEAME.
	    The Reg Methods in row (e) are the word embedding label smoothing and the projection matrix regularization methods applied jointly.
	    Model averaging was applied to rows (f) and (i).
	}
	\label{tab:all}
	\centering
    \vspace{10pt}

	\begin{tabular}{lccc}
        \toprule
        \textbf{Method} & \small{\textbf{devman}} & \small{\textbf{devsge}} & \small{\textbf{RTF}} \\
        \midrule
        \multicolumn{4}{c}{\textit{(I) Non-autoregressive}} \\
        \midrule
        (a) CTC & 24.2 & 35.0 & 0.01 \\
        (b) Mask-CTC & 16.5 & 24.4 & 0.02 \\
        (c) + M2M (w/o~Pinyin) & 16.6 & 24.4 & 0.02 \\
        (d) + P2M (w/~~Pinyin) & 16.3 & 24.0 & 0.02 \\
        (e)~~~+ Reg Methods & 16.0 & 24.1 & 0.02 \\
        (f)~~~~~~+ Model Avg & \textbf{15.3} & \textbf{22.3} & 0.02 \\
        \midrule
        \multicolumn{4}{c}{\textit{(II) Autoregressive}} \\
        \midrule
        (g) Transformer-T \cite{dalmia2021transformer} & 18.5 & 26.3 & $-$ \\
        (h) Multi-Enc-Dec \cite{zhou2020multi} & 16.7 & 23.1 & $-$ \\
        (i) Conformer (Ours) & \textbf{14.3} & \textbf{20.6} & 1.03 \\
		\bottomrule
	\end{tabular}
\end{table}

This section investigated the effectiveness of the proposed model and regularization methods using all training data in SEAME.
We chose models with the lowest validation losses for evaluation, and the results are listed in Sec. (I) of Table \ref{tab:all}.

We set three baselines: conformer CTC (row (a)), Mask-CTC (row (b)), and the proposed architecture but used Chinese characters instead of Pinyin tokens as the intermediate, denoted as the Mandarin-to-Mandarin (M2M) decoder (row (c)).
First, the Mask-CTC architecture obtained better performance than the original CTC model (rows (b) v.s. (a)), indicating that leveraging the CMLM decoder was beneficial for CS speech recognition.
Next, adding the M2M decoder to the Mask-CTC framework slightly damaged the ASR performance (rows (c) v.s. (b)).
However, switching the encoder's target to Pinyin improved the ASR performance (rows (d) v.s. (b)), indicating that introducing Pinyin as the intermediate representation helped recognition of Chinese characters.
Moreover, the regularization methods achieved better results on devman and obtained comparable results on devsge (rows (e) v.s. (d)).
To decrease the variance of the model's prediction, we selected the top five checkpoints with higher validation accuracy for averaging \cite{karita2019comparative}, resulting in the best NAR ASR performance (row (f)).

To highlight the benefits of using NAR ASR models, we listed the performance of AR ASR models in Sec. (II) of Table~\ref{tab:all} for comparison.
Our AR baseline incorporated an RNN-LM by shallow fusion with a beam size of 10.
The model surpassed the previous SOTA Transformer-transducer \cite{dalmia2021transformer} and Multi-encoder-decoder \cite{zhou2020multi} (rows (i) v.s. (g)(h)), providing a solid reference.
The best NAR model offered good performance with a small gap between the best AR model (rows (f) v.s. (i)) but with a significantly 50$\times$ speedup (the RTF column), implying the NAR model could recognize CS speech while possessing fast inference speed.
Overall, we have shown that NAR ASR models incorporated with the proposed methods achieved exciting results.

\subsection{Regularization Methods for Low-resource Setting}
\label{subsec:ablation}
\begin{table}[t]
	\caption{
	    Comparison of regularization methods in low-resource scenario in TER(\%).
	    EmbLS and MatReg respectively denote word embedding label smoothing and projection matrix regularization.
	    The overall TER is in the \textbf{All} column.}
	\label{tab:ablation}
	\centering
	\vspace{10pt}
	\begin{tabular}{ccc|cc|c}
        \toprule
        & \textbf{EmbLS} & \textbf{MatReg} & \textbf{devman} & \textbf{devsge} & \textbf{All} \\
        \midrule
        (a) & \XSolidBrush & \XSolidBrush & 42.2 & 50.7 & 45.3 \\
        \midrule
        (b) & \CheckmarkBold & \XSolidBrush & 42.0 & 50.1 & 44.9\\
        (c) & \XSolidBrush & \CheckmarkBold & 42.2 & 50.4 & 45.2\\
        (d) & \CheckmarkBold & \CheckmarkBold & 41.7 & 50.2 & 44.8\\
		\bottomrule
	\end{tabular}
\end{table}

To verify the efficacy of the proposed regularization methods, we conducted ablation studies under the low-resource scenario with only 10\% of training data. \footnote{Due to space limitation, we only show results in low-resource settings.}
The performance of the baseline model is shown in row (a) of Table~\ref{tab:ablation}.
Improvement was brought by only applying the word embedding label smoothing (rows (b) v.s. (a)), showing that leveraging textual knowledge from additional text data was beneficial under the low-resource scenario.
With the proposed projection matrix regularization applied, the model obtained less improvements (rows (c) v.s. (a)).
Finally, applying both methods (row (d)) achieved the best performance.
We showed that the projection matrix regularization technique benefited from the label smoothing method and filled the gap between the encoder and decoders.
The ablation study showed that both methods were compatible with each other and improved ASR performance.

\subsection{Error Analyses}
\begin{table*}[t]
    \caption{
        A more detailed error analysis of different methods on the combination of devman and devsge evaluation sets.
        All values are in percentage (\%).
        The overall TER are divided into (i) all tokens, (ii) Mandarin characters, (iii) Pinyin tokens, and (iv) English words.
        The \textbf{Sub}, \textbf{Del}, and \textbf{Ins} columns respectively indicate substitution, deletion, and insertion.
        The deletion rates are also divided into (vi) all tokens, (vii) Mandarin characters, and (viii) English words.
    }
    \label{tab:error}
    \centering
    \vspace{6pt}
    \begin{tabular}{l|cccc|c|ccc|c}
        \toprule
        \multirow{2}{*}{\textbf{Method}}
         & \multicolumn{4}{c|}{\textbf{TER}} & \textbf{Sub} & \multicolumn{3}{c|}{\textbf{Del}} & \textbf{Ins} \\
        \cmidrule{2-10}
         & \small (i) All & \small (ii) Man & \small (iii) Pinyin & \small (iv) Eng & \small (v) All & \small (vi) All & \small (vii) Man & \small (viii) Eng & \small (ix) All \\
        \midrule
        (a) CTC & 28.1 & 13.2 & 11.8 & 14.9 & 17.4 & 8.9 & 4.5 & 4.3 & 1.8 \\
        (b) Mask-CTC & 19.4 & 9.0 & 8.0 & 10.4 & 12.5 & 5.1 & 2.8 & 2.3 & 1.8  \\
        (c) + M2M    & 19.4 & 9.0 & 7.9 & 10.5 & 12.7 & 4.7$^{\dagger\ast}$ & 2.5$^{\dagger\ast}$ & 2.2 & 1.9$^{\ast}$ \\
        (d) + P2M    & 19.1 & 8.9 & 7.7 & 10.2 & 12.7 & 4.5$^{\dagger\ast}$ & 2.5$^{\dagger\ast}$ & 2.1$^{\dagger\ast}$ & 1.9 \\
        (e)~~~+ Reg Methods & 18.9 & 8.8 & 7.6 & 10.1 & 12.7 & 4.3$^{\dagger\ast}$ & 2.3$^{\dagger\ast}$ & 2.0$^{\dagger\ast}$ & 1.9 \\
        \midrule
        (f)~~~~~~+ Model Avg & 17.8 & 8.4 & 7.3 & 9.5 & 11.8 & 4.2 & 2.4 & 1.8 & 1.8 \\
        (g) AR & 16.6$^{\ddagger\diamond}$ & 7.9$^{\diamond}$ & 6.9$^{\diamond}$ & 8.6$^{\ddagger\diamond}$ & 11.0$^{\ddagger\diamond}$ & 3.8$^{\ddagger\diamond}$ & 2.2$^\ddagger$ & 1.6$^{\ddagger\diamond}$ & 1.7$^{\diamond}$ \\
        \bottomrule
    \end{tabular}
    \vspace{-2pt}
    \begin{flushleft}
    $^\dagger$ Passed t-test with $p < $ 0.01 compared with row (b).
    $^\ddagger$ Passed t-test with $p < $ 0.01 compared with row (f).\\
    $^\ast$ Passed McNemar test with $p < $ 0.05 compared with row (b).
    $^\diamond$ Passed McNemar test with $p < $ 0.05 compared with row (f).
    \end{flushleft}
    \vspace{-14pt}
\end{table*}

This section analyzed each method's errors and showed how the proposed methods decreased the error rates.
The results over devman + devsge are shown in Table \ref{tab:error}.
We analyzed the source of TERs, including substitution, deletion, and insertion.
We provided the Pinyin error rates (PER, column (iii)), by mapping all Chinese characters to their corresponding Pinyin tokens, and evaluated TER at Pinyin level.
PER shows the models' ability to produce Chinese characters with correct pronunciations.
We conducted t-test and McNemar test to compare the significance of improvements.

We first inspected the methods without model averaging in rows (a) to (e).
Most results in columns (i) to (iv) were very similar to Table \ref{tab:all}, but more information could be extracted with separated TERs of Mandarin and English symbols.
First, the Mandarin and English errors contributed similarly to the total TERs.
With the proposed P2M decoder and regularization methods respectively added, both Mandarin and English TERs were reduced (rows (d)(e) v.s. (b)), showing that our approaches benefited the two languages simultaneously.
Another observation was that the ratio between PERs and Mandarin TERs ((iii) $\div$ (ii)) gradually dropped from 89.5\% (row (a)) to 85.9\% (row (d)) when adding the proposed methods, indicating that the models tend to predict Chinese characters with correct pronunciation with our methods.

\begin{figure}[t]
    \centering
    \includegraphics[width=\linewidth]{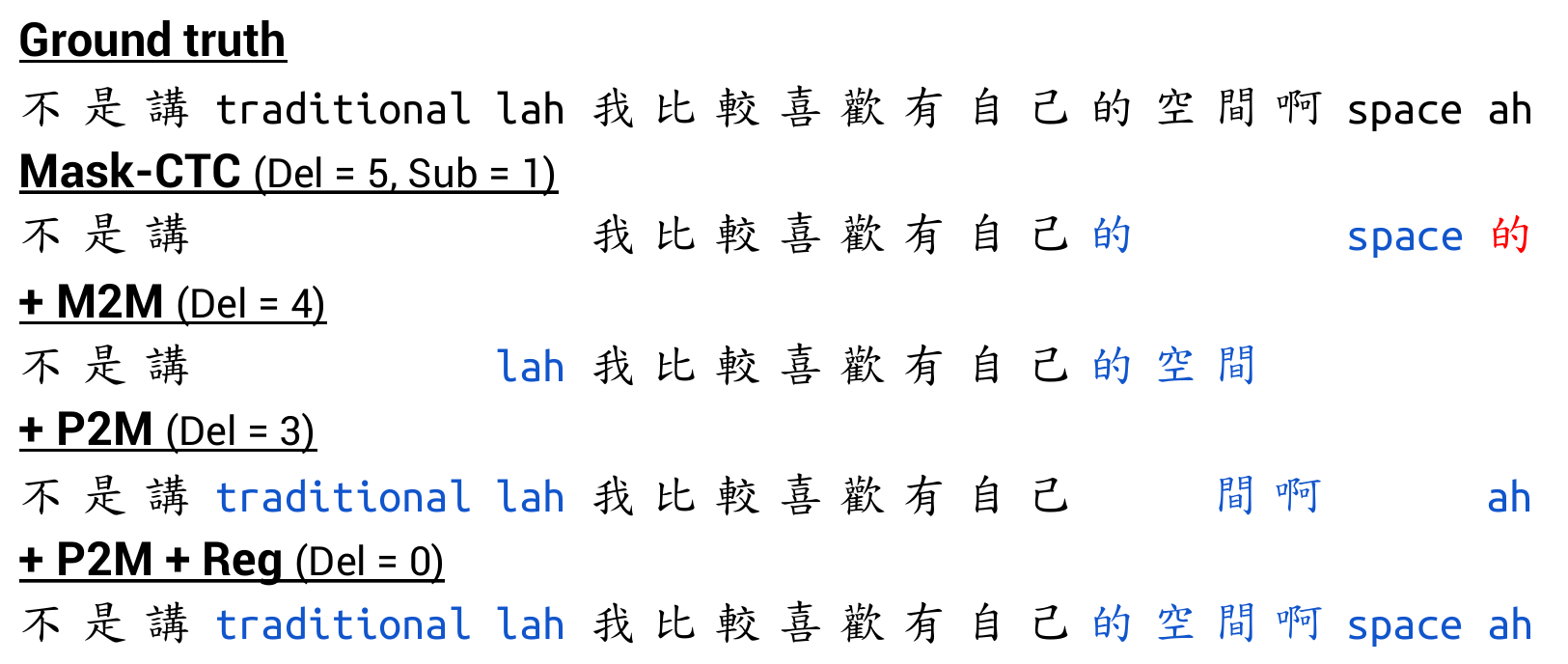}
    \vspace{-20pt}
    \caption{
        Samples of different ASR models' predictions of the same utterance from devman.
        Del and Sub here indicate the number of required deletions and substitutions, respectively.
        Fewer deletions have to be made for the proposed methods.
    }
    \label{fig:samples}
    \vspace{-5pt}
\end{figure}
Next, we investigated the contributions of substitution, deletion, and insertion to the overall TER (columns (v) to (ix)).
We found that the number of substitutions and insertions were almost the same among different methods (columns (v) and (ix)), but deletions had significant improvements (columns (vi) to (viii)).
Results indicated that our approaches made the ASR output sequence length closer to the ground truth.
An actual sample is shown in Fig. \ref{fig:samples}, Mask-CTC prone to predict shorter sentences on the SEAME corpus, but our methods provided more accurate predictions in both Chinese characters and English words, resulting in lower deletion rates.
Moreover, the proposed P2M decoder passed t-test and McNemar tests in English deletion while the M2M decoder failed (column (viii), rows (d) v.s. (c)), showing that using Pinyin as the intermediate benefited English word prediction.
Replacing Chinese characters with Pinyin symbols forced the encoder to learn acoustic modeling better; thus, it predicted more accurate sequence lengths rather than struggling to predict correct Chinese characters directly.

Finally, we compared the differences between our best NAR model and the AR baseline (rows (f)(g)).
In most error terms, the AR baseline surpassed the best NAR model with high confidence, showing the large gaps between the NAR and AR ASR models.
Therefore, we should pay more attention to these gaps, including English TER, substitution rates, and deletion rates.
In contrast, the Mandarin and Pinyin TERs and the insertion rate failed to pass the t-test, showing that the NAR model's performance was similar to the AR model in these metrics.
These results provided some insights to improving NAR CS speech recognition models.
For instance, to decrease the deletion rate of a NAR ASR, more efforts can be made in the decoding stage to predict more precise length of the output \cite{higuchi2021improved-maskctc}.
Overall, the results have shown the proposed methods offered powerful performance compared to the AR baseline, while remains some space to improve.

\section{Conclusion}
\label{sec:conclusion}

This paper introduces a novel non-autoregressive ASR framework by adding a Pinyin-to-Mandarin decoder in the Mask-CTC ASR to solve the Mandarin-English code-switching speech recognition problem.
We also propose word embedding label smoothing for including contextual information to conditional masked language model and projection matrix regularization method to bridge the gap between the encoder and decoders.
We demonstrated the effectiveness of our methods with exciting performance on the SEAME corpus.
The new ASR framework and regularization methods have the potential to improve various speech recognition scenarios.

\bibliographystyle{IEEEbib}
\bibliography{refs,cs}

\end{document}